%% file: ICIP2014.tex
\documentclass{article}

\usepackage{spconf,amsmath,amsfonts,amssymb,amsbsy}
\usepackage{amsthm}
\usepackage{graphicx}
\usepackage{float}
\usepackage{verbatim}
\usepackage{pslatex}
\usepackage{cite,url,bm}
\usepackage{algorithm}
\usepackage{algorithmic}
\usepackage{epsf,subfigure,psfig}
\usepackage{latexsym,amsmath,epsfig}
\usepackage{multirow}
\renewcommand{\algorithmicrequire}{\textbf{Input:}}
\renewcommand{\algorithmicensure}{\textbf{Output:}}
\newcommand{\vect}[1]{\pmb{#1}}
\newcommand{\mat}[1]{\pmb{#1}}

\def\ben{\begin{equation*}}
\def\een{\end{equation*}}
\def\be{\begin{equation}}
\def\ee{\end{equation}}
\def\beaa{\begin{eqnarray*}}
\def\eeaa{\end{eqnarray*}}
\def\bea{\begin{eqnarray}}
\def\eea{\end{eqnarray}}

\begin{document}
\title{Multi-Task Image Classification Via Collaborative, Hierarchical\\ Spike-And-Slab Priors}
 
\twoauthors
  {~~Hojjat S. Mousavi, U. Srinivas, V. Monga }
	{Dept. of Electrical Engineering \\
	The Pennsylvania State University \thanks{Copyright \copyright  2010 IEEE. Personal use of this material is permitted. However, permission to use this material for any other purposes must be obtained from the IEEE by sending a request to  pubs-permissions@ieee.org.}}
  {Yuanming Suo, M. Dao, T. D. Tran }
	{Dept. of Electrical and Computer Engineering\\
	The Johns Hopkins University}

\maketitle

\begin{abstract}
Promising results have been achieved in image classification problems by exploiting the discriminative power of sparse representations for classification (SRC). Recently, it has been shown that the use of \emph{class-specific} spike-and-slab priors in conjunction with the class-specific dictionaries from SRC is particularly effective in low training scenarios. As a logical extension, we build on this framework for multitask scenarios, wherein multiple representations of the same physical phenomena are available. We experimentally demonstrate the benefits of mining joint information from different camera views for multi-view face recognition.
\end{abstract}
\input{section1_v3} \vspace{-4mm}
\input{section2_v3}
\input{section3_v3}
\input{section4_v3}

% To start a new column (but not a new page) and help balance the last-page
% column length use \vfill\pagebreak.
% -------------------------------------------------------------------------
%\vfill
%\pagebreak

\begingroup
\fontsize{9pt}{9pt}\selectfont
\bibliographystyle{IEEEtran}
\bibliography{IEEEabrv,ICIP2014}
\endgroup
%%%%%%%%%%%%%%%%%%%%%%%%%%%%%%%%%%%%%%%%%%%%%%%%%%%%%%%%%%%%%%%%%%%%%%%%%%%%%%%%
%%%%%%%%%%%%%%%%%%%%%%%%%%%%%%%%%%%%%%%%%%%%%%%%%%%%%%%%%%%%%%%%%%%%%%%%%%%%%%%%
\end{document}

%% file: section1_v3.tex
\section{Introduction}
\label{sec:intro}
%\subsection{Sparse Representation-based Classification }
%\label{subsec:SRC}
Image classification is an important problem that has been studied for over three decades. Practical applications  span a wide variety of areas such as general texture or object categorization\cite{Zhang:SVMKNN_CVPR06, Yuan:VisualClassificationMultitask_TIP12} , face recognition \cite{Zou:ComparativeStudyFaceRecognition_TIP07, Fukui:MSM_Springer05} and automatic target recognition in hyperspectral or radar imagery \cite{Ren:ATR_Hyperspectral_AeroElecSys03, Zhao:SVM_SAR_ATR_AeroElecSys01}. Various methods of feature extraction (\cite{Haralick:TextureFeatureImageClassification_TSMC73, Lazebnik:BagOfFeatures_CVPR06} for example) as well as classifiers \cite{Zhang:SVMKNN_CVPR06,Rande:TexttureClassificationReview_PAMI99} have been investigated.

The advent of of compressive sensing (CS) \cite{Donoho:CS_InfoTheory06} has inspired research in the direction of applying the central analytical formulation of CS to classification problems. Sparse representation-based classification (SRC)\cite{Wright:SRC_tpami09} is arguably the most well-known such tool that has demonstrated robust performance even in the presence of high pixel distortion, occlusion or noise. Extensions of SRC have been proposed along two lines of thought: (i) by adding regularizer and priors which prevent overfitting issue by introducing additional information to the problem and (ii) by exploiting joint information and complementary data in multitask cases. We are also moving along this direction to use the advantages of using priors as well as joint information.

\noindent \textbf{Motivation and Contribution:}
Advances in sensing technology have facilitated the easy acquisition of multiple different measurements of the same underlying physical phenomena. Often there is complimentary information embedded in these different measurements which can be exploited for improved performance. For example, in face recognition or action recognition we could have different views of a person's face captured under different illumination conditions or with different facial postures \cite{Zhang:JointDynamicSparseFaceRecognition_pattern12, Yuan:VisualClassificationMultitask_TIP12, Gross:MultiPIE, Guha:learningPAMI12, Bahrampour:qualityCVPR14}. In automatic target recognition, multiple SAR (synthetic aperture radar) views are acquired \cite{Zhang:MultiViewATR_taes12}. The use of complimentary information from different color image channels in medical imaging has been demonstrated in \cite{srinivas:isbi13}. In border security applications, multi-modal sensor data such as voice sensor measurements, infrared images and seismic measurements are fused \cite{Nasrabadi:MultiSensorFusion_Fusion11} for activity classification tasks. The prevalence of such a rich variety of applications where multi-sensor information manifests in different ways is a key motivation for our contribution in this paper. Specifically, we extend recent work in class-specific sparse prior-based classification by Srinivas \emph{et al.} \cite{Srinivas:StructuredSparsePriors_ICIP13} to a multitask framework.

We extend the Bayesian framework in \cite{Srinivas:StructuredSparsePriors_ICIP13} in a hierarchical manner in order to capture joint information across multiple tasks (or measurements). As observed in \cite{Srinivas:StructuredSparsePriors_ICIP13}, an important challenge is to develop a framework based on spike-and-slab priors that can capture a general notion of signal sparsity while also leading to tractable optimization problems. Our contribution successfully addresses both these issues via a generalized collaborative Bayesian hierarchical method. Expectedly it results in a hard non-convex optimization problem. We propose an efficient solution using Monte Carlo Markov Chain (MCMC) method which results in practical benefits as demonstrated in Section \ref{sec:Results}.
%The advantage of using this hierarchical structure is that instead of forcing coefficients to be zero or non-zero, a probabilistic method can control the structure of coefficient matrix through learning $\kappa$. This hierarchical method benefits from its probabilistic nature, however, there are some simplifying and also limiting assumptions. For example, it is assumed that there is only one $\kappa$ per class not per coefficient. This assumption leads to a simpler and tractable optimization problem but at the same time it is not a general framework that can capture sparsity and structure on the level of coefficients. In this paper, we aim to resolve this issue and propose a general framework that can capture more general notions of sparsity and structure in the coefficient vector. We addressed this issue by assuming different $\kappa$ values for each coefficient which leads to a harder optimization problem, however, an efficient solution is proposed.\\
%\begin{itemize}

%% file: section2_v3.tex
\section{Sparse Representation-based Classification (SRC)}
\label{sec:review}

The aim of CS is essentially to recover a higher dimensional compressible (sparse) signal $\vect x \in \mathbb{R}^n$ from a set of lower dimensional linear measurements $\vect y \in \mathbb{R}^m$ of the form $\vect y = \mat A \vect x$. It is shown that $\vect x$ is recoverable by solving the following underdetermined ($m \ll n$) optimization problem\cite{Candes:CS_RobustUncertaintyPrinciples}:
\be
    \min_{\vect x} ~||\vect x||_0 \textit{~~subject to~~}  \vect y = \mat A \vect x,
\label{Eq:CS_main1}
\ee
where $||\vect x||_0$ is basically the number of non-zero elements of $\vect x$. It is known that (\ref{Eq:CS_main1}) is an NP-hard problem but can be solved by relaxing the non-convex term \cite{Donoho:CS_Relaxed_cpam06} and solve the following optimization problem in presence of noise. %In addition, in real world applications, where signals are noisy, the equality constraint is changed to an inequality constraint to tolerate a fixed level of noise $\varepsilon$ \cite{Tibshirani:RegressionShrinkage}:
%\be
%    \min_{\vect x} ~||\vect x||_1 \textit{~~subject to~~}  \vect y = \mat A \vect x,
%\label{Eq:CS_Relaxed}
%\ee
\be
    \min_{\vect x} ~||\vect x||_1 \textit{~~subject to~~}  ||\vect y - \mat A \vect x||_2< \varepsilon.
\label{Eq:CS_Relaxed_Noisy}
\ee
Recently, \emph{Wright et al.} proposed SRC \cite{Wright:SRC_tpami09} for face recognition by designing an over-complete \emph{class-specific} dictionary $\mat A$ and employing CS framework. For a multi-class problem, the dictionary matrix $\mat A$ is built using training images from all classes as $\mat A = [\mat A_1~ \ldots ~\mat A_C]$  where each column of $\mat A_{C_r}$ is a vectorized training image (dictionary atom) from class $C_r$.

The minimization problem in \eqref{Eq:CS_Relaxed_Noisy} is equivalent to maximizing probability of observing the sparse vector $\vect x$ given $\vect y$ assuming $\vect x$ has an i.i.d Laplacian distribution in a Bayesian framework\cite{Babacan_BayesianCSLaplacePriors_TIP10}. Therefore, sparsity can be interpreted as a prior on the coefficient vector which enhances signal comprehension by incorporating contextual information and signal structure. Structure and sparsity both can be enforced by introducing probabilistic priors, $f(\vect x)$, on the coefficient vector and solving the following optimization problem \cite{Cevher_LearningCompressiblePriors_NIPS09}:
\be
    \max_{\vect x} f(\vect x) \textit{~~subject to~~}  ||\vect y - \mat A \vect x||_2< \varepsilon.
\label{Eq:PriorOpt}
\ee
%The most challenging part is to find a prior which can both capture sparsity and the structure of coefficient vector. In fact a very well-suited structured prior for sparsity is the spike-and-slab prior \cite{Ishwaran_SpikeSlab_AnnStat05,George_VariableSelectionGibbsSampling_StatAssoc93,Carvalho_SparseFactorModeling_StatAssoc08,Chipman_BaysianVariableSelection_Stat96} which is known to be the gold standard for sparse inference in Bayesian framework \cite{TitsiasL_SpikeSlabGoldStandard_NIPS11}. Using spike-and-slab priors, each coefficient of the sparse vector is modeled as mixture of spike and slab components:
%\be
%    x_i \sim (1-\gamma_i) \delta_0 + \gamma_i f(x_i)
%\label{Eq:SpikeAndSlabIntro}
%\ee
%where $\delta_0 $ is a point mass concentrated at zero (known as "spike") and $f$ is the distribution of nonzero coefficients of sparse vector also known as "slab". Note that the choice of $\gamma_i\in[0,1]$ can enforce both structure and sparsity of signal.\\
Recently, Srinivas \emph{et al.} proposed the use of spike-and-slab priors under a Bayesian framework for image classification purposes \cite{Srinivas:StructuredSparsePriors_ICIP13}. They proposed one class-conditional probability distribution function (pdf) per class represented by $f_{C_1} $ and $f_{C_2}$ with the same class-specific dictionary $\mat{A}$ as before. Given a test vector $\vect y$ class assignment is done  by solving the following constrained likelihood maximization problem per class where $f_{C_r}$'s are learned separately using spike-and-slab priors:
\be
    \hat{\vect x}_{C_r}  = \arg\max_{\vect x} f_{C_r}(\vect x) \textit{~~subject to~~}  ||\vect y - \mat A \vect x||_2< \varepsilon.
\label{Eq:MaxPrior}
\ee
\be
    \mbox{Class}(\vect y) = \arg\max_{r\in\lbrace 1,2 \rbrace } f_{C_r}(\hat{\vect x}_C{_r})
\label{eq:ClassAssign1}
\ee
Inspired by \cite{Yen_VAriableSelectionSpikeSlab_Stat11} they developed a Bayesian framework for classification and considered a linear model $\vect y = \mat A \vect x +\vect n$ in each class where $\vect y \in \mathbb{R}^m$, $\vect x \in \mathbb{R}^n$, $\mat A \in \mathbb{R}^{m\times n}$ and $\vect n$ is the inherent Gaussian noise (we drop the class indices for notational simplicity). Using this set-up the underlying Bayesian framework is as follows:
\small
\bea
    \vect y | \mat A, \vect x, \vect \gamma, \sigma^2 & \sim &~\mathcal{N} (\mat A \vect x, \sigma^2\mat I )
\label{Eq:BaysianFramework1_y}\\
    x_i | \sigma^2, \gamma_i, \lambda & \sim & \gamma_i\mathcal{N}(0,\sigma^2\lambda^{-1}) + (1-\gamma_i)\delta_0,~i = 1,\ldots,n
\label{Eq:BaysianFramework1_x}\\
%    \sigma^2 | \tau_1,\tau_2 & \sim &~\mbox{Inverse-Gamma}(\tau_1,\tau_2) \\
    \gamma_i | \kappa & \sim &~\mbox{\small{Bernoulli}}(\kappa),~i = 1,\ldots,n.
\label{Eq:BaysianFramework1_gamma}
\eea
\normalsize
where $\mathcal{N}(.)$ represents the normal distribution and \eqref{Eq:BaysianFramework1_x} is modeling each $x_i$ with a spike-and-slab prior which is a very well-suited structured prior for capturing sparsity \cite{Ishwaran_SpikeSlab_AnnStat05, George_VariableSelectionGibbsSampling_StatAssoc93, TitsiasL_SpikeSlabGoldStandard_NIPS11}. $\delta_0 $ is a point mass concentrated at zero (known as "spike") and the other term is the distribution of nonzero coefficients of sparse vector also known as "slab". In this framework $\gamma_i$'s are assumed to have binary values, either 1 (if $x_i=1$) or zero (if $x_i \ne 0$) in which case they become indicator variables for each element of $\vect x$. It is clear that $\vect \gamma$ can control the sparsity level of the signal and at the same time enforce a specific structure. As a result, $\kappa$, which is the probability of each coefficient to be nonzero, plays a key role in identifying the structure. In the next section, we will show how a smart choice of $\kappa$ can lead to a framework that is more general and able to capture different sparse structures in the coefficient vector.\\

%% file: section3_v3.tex
\section{Multi-Task Image Classification Via Collaborative, Hierarchical Spike-And-Slab Priors}
\label{sec:CollaborativePriors}
\subsection{Bayesian Framework}
\label{subsec:OptProb}
\begin{figure}
    \centering
    \includegraphics[scale=0.38]{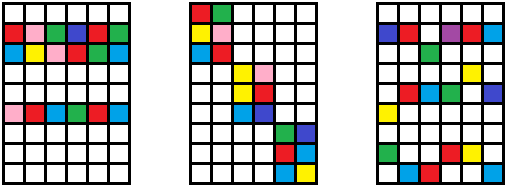}
    \caption{Row, block and dynamic sparsity from left to right.}
    \label{Fig:Notions}
\end{figure}

If there are multiple measurements $\vect{y_1},\ldots,\vect{y_T}$ of the same signal, we now have a sparse coefficient matrix $\mat X$ as follows:
\be
\mat Y := [\vect{y_1} ~ \ldots ~ \vect {y_T}] = \mat A[\vect x_1 ~ \ldots ~ \vect x_T] = \mat A \mat X.
\ee
Depending on the application and type of measurement, different notions of matrix sparsity can occur with respect to the coefficient matrix $\mat X$. As illustrated in Fig. \ref{Fig:Notions}, in some applications such as hyperspectral classification, row sparsity - entire rows with all zero or all non-zero coefficients - emerges naturally in matrix $\mat X$, whereas in other applications, block sparsity or joint dynamic sparsity are more appropriate. This sparsity pattern is an inherent feature of the application and the specific way in which the multi-task dictionary has been designed. Our contribution is a  framework that is applicable in a wide variety of applications by capturing a general notion of the sparse structure in the coefficient matrix.

Assume we have $T$ observations (tasks) from the same class and  put them together to form a matrix $\mat Y$. We desire to find a sparse matrix $\mat X$ such that we can guarantee that the error between test matrix and its reconstructed version is small and at the same time, matrix $\mat X$ has a sparse structure.\\
We modified \eqref{Eq:BaysianFramework1_y}-\eqref{Eq:BaysianFramework1_gamma} in order to generalize the Bayesian framework to multitask case. This modified version should be able to model the behaviour of $\mat X$ and induce the desired structure and sparsity in $\mat X$. Generalized Bayesian framework for multitask case is as follows:

\small \vspace{-0.12in}
\bea
    \mat Y | \mat A, \mat X, \mat \Gamma, \sigma_n^2 & \sim &~ \prod_{t=1}^{T} ~\mathcal{N} \left(\mat A \vect x_t, \sigma_n^2\mat I \right)
\label{Eq:BaysianFramework2_Y}\\
    \mat X | \sigma^2, \mat\Gamma, \lambda & \sim & \prod_{t=1}^{T} \prod_{i=1}^{n} ~\gamma_{t_i} \mathcal{N}(0,\sigma^2\lambda^{-1}) + (1-\gamma_{t_i}) \delta_0
\label{Eq:BaysianFramework2_X}\\
    %\sigma^2 | \tau_1,\tau_2 & \sim &~\mbox{Inverse-Gamma}(\tau_1,\tau_2) \\
    \mat\Gamma | \mat K & \sim & \prod_{t=1}^{T} \prod_{i=1}^{n} ~\mbox{Bernoulli}(\kappa_{t_i}).
\label{Eq:BaysianFramework2_Kappa}
\eea
\normalsize
Where $\vect x_t$ is the $t^{th}$ column of the matrix $\mat X$, $\mat K = \{\kappa_{t_i} \}_{t,i}$ and $\mat \Gamma = \{\gamma_{t_i} \}_{t,i}$ for $t=1...T$ and $i=1...n$ \\
\textbf{Remark:} Note that in contrast to \eqref{Eq:BaysianFramework1_gamma} as was proposed in \cite{Srinivas:StructuredSparsePriors_ICIP13}, $\kappa$ is assumed to take different values for each task and each coefficient. With this assumption our framework can capture more general notions of structure and sparsity in the matrix $\mat X $. This is one of our central analytical contributions in this paper, in contrast with methods with relaxed and simplified assumptions \cite{Srinivas:StructuredSparsePriors_ICIP13 ,Suo:HierarchicalSpikeSlabPriors_ICASSP13}.

The benefit of using the Bayesian approach is that it can alleviate the burden on requirement of abundant training. One of the central assumptions in SRC is existence of sufficient training (overcomplete dictionary $\mat A$) whereas proposed Bayesian approach can handle scenarios that lack the number of training. This is enabled by use of class-specific priors that can offer more discriminability in the dictionary.

To perform the Bayesian inference we followed a similar procedure as in \cite{Srinivas:StructuredSparsePriors_ICIP13} to obtain the joint posterior density function for MAP estimation.
%\small
%\be
%    f(\mat X,\mat \Gamma, |\mat A,\mat Y,\lambda, \mat K) \propto f(\mat Y |\mat A,\mat X,\mat \Gamma,\sigma^2)f(\mat X|\mat \Gamma,\sigma^2,\lambda) f(\mat \Gamma | \mat K).
%\label{eq:joint-posterior}
%\ee
%\normalsize
%The optimal $\mat X^\ast, \mat \Gamma^\ast $ are obtained by MAP estimation as:
%\small
%\be
%    (\mat X^\ast, \mat\Gamma^\ast ) = \arg \min_{\mat X,\mat \Gamma, } \left\{-2\log f(\mat X,\mat \Gamma, |\mat A,\mat Y,\lambda, \mat K)\right\}.
%\label{eq:map-estimate}
%\ee
%\normalsize
Here we only present the resulting optimization problem; the detailed discussion of the optimization problem is available online in a technical report at \cite{Mousavi_OptProb:Techrep13}.
It must be noted that we have a different framework for each class and as a result, we get $C$ different optimization problems to find $\mat X_{C_r}^\ast$ and $\mat\Gamma_{C_r}^\ast $:
\bea
    \arg\min_{\mat X, \mat \Gamma}~ \frac{\sigma^2}{\sigma_n^2} ||\mat Y - \mat A\mat X ||_F^2 +\lambda ||\mat X||_F^2 +
     \sum_{t=1}^{T} \sum_{i=1}^{n}   \gamma_{t_i} \rho_{t_i}
    \label{Eq:FinalOptProb}
\eea
where $\rho_{t_i} = \sigma^2   \log \big(\frac{2\pi\sigma^2 (1-\kappa_{t_i})^2}{\lambda\kappa_{t_i}^2} \big)$. First term in \eqref{Eq:FinalOptProb} is basically trying to minimize the reconstruction error, whereas the second and third terms are jointly trying to keep the coefficient matrix smooth and sparse.\\
\textbf{Remark:} Solution to this optimization problem has not been addressed yet in the literature but its relaxations reduce to well-known problems in compressive sensing and statistics such as LASSO, Elastic Net, etc \cite{Tibshirani:RegressionShrinkage, Sprechmann:CHilasso_TSP11, Suo:HierarchicalSpikeSlabPriors_ICASSP13}.\\
Finally, after solving each optimization problem per class for a given test matrix $\mat Y$, the class assignment will be as follows.
\bea
    \mbox{Class}(\mat Y) &=& \arg\min_{r\in\lbrace 1,...,C \rbrace } L_r(\mat X^\ast_{C_r} ; \mat \Gamma_{C_r}^\ast )
\label{Eq:ClassAssignCollaborative1}
    %\mbox{Class}(\mat Y) &=& \arg\min_{i\in\lbrace 1,...,C \rbrace } ||\mat Y - \mat A \mat X_{C_i}^\ast||_F
%\label{Eq:ClassAssignCollaborative2}
\eea
\subsection{Solution to Optimization Problem }
\label{subsec:OptSolu}
In this section we provide an efficient solution to the resulting optimization problem in \eqref{Eq:FinalOptProb}. First, we rewrite the problem as follows:
\bea
     \arg\min_{\mat X, \mat \Gamma} ~~ \sum_{t=1}^{T} \bigg( \frac{\sigma^2}{\sigma_n^2}||\vect y_t  - \mat A\vect x_t ||_2^2 +
     \lambda ||\vect x_t||_2^2 + \sum_{i=1}^{n}  \gamma_{t_i} \rho_{t_i} \bigg)
\label{Eq:FinalOptProb2}
\eea
As it can be seen, it consists of $T$ different independent optimization problems which can be solved individually rather than solving a complex matrix optimization problem. Henceforth, for each class we should follow the following procedure:
\begin{itemize}
    \item $\underset{\vect x_t, \vect \gamma_t}{\textit{minimize~}} \frac{\sigma^2}{\sigma_n^2}||\vect y_t  - \mat A\vect x_t ||_2^2 + \lambda ||\vect x_t||_2^2 + \sum_{i=1}^{n}  \gamma_{t_i} \rho_{t_i} $, $\forall t$ \vspace{-0.1in}
    \item Put $\vect x_t$'s and $\vect \gamma_t$'s together to form $\mat X$ and $\mat \Gamma$, respectively.
\end{itemize}\vspace{-0.06in}
From now onward we only talk about the optimization problem in class $C_r$ and task $t$. However, it should be solved for each $C_r,~r =1,...,C$ and $t, ~t=1,...,T$, separately. For the sake of simplicity, we drop class and task indices and use $\vect y$, $\vect x $ and $\vect \gamma $ instead of $\vect y_t$, $\vect x_{t}$ and $\vect \gamma_{t}$, respectively.

Note that the above optimization problem is a hard non-convex problem. We propose an efficient solution using MCMC method.
%One of the advantages of introducing a hierarchical model in \eqref{Eq:BaysianFramework2_Y}-\eqref{Eq:BaysianFramework2_Kappa} is that we can obtain the marginal distribution $f(\mat \Gamma|\mat Y) \propto f(\mat Y |\mat \Gamma)f(\mat \Gamma)$. According to \cite{George_VariableSelectionGibbsSampling_StatAssoc93}, $f(\mat \Gamma|\mat Y)$ provides a posterior probability that can be used to select more promising atoms that contribute in reconstructing $\mat Y$. Henceforth, instead of calculating all $2^n$ possible posteriors, we can use Gibbs sampler in order to extract promising atoms that contribute more in reconstruction. In addition, note that finding the optimum $\mat \Gamma^\ast$ is equivalent to find the most important atoms in the dictionary $\mat A$ that have the most contribution in reconstructing $\mat X$. \\
One of the advantages of introducing a hierarchical model in \eqref{Eq:BaysianFramework2_Y}-\eqref{Eq:BaysianFramework2_Kappa} is that we can obtain the marginal distribution $f(\vect \gamma|\vect y) \propto f(\vect y |\vect \gamma)f(\vect \gamma)$. According to \cite{George_VariableSelectionGibbsSampling_StatAssoc93}, $f(\vect \gamma|\vect y)$ provides a posterior probability that can be used to select promising atoms that contribute more in reconstructing $\vect y$. Moreover, finding the optimum $\vect  \gamma^\ast$ is also equivalent to finding the most prominent atoms in the dictionary $\mat A$. Henceforth, we propose the following scheme to solve the optimization problem: first, we find dictionary atoms that contribute in reconstructing the test vector $\vect y$. Then we find the value of contribution for each atom we found previously.\\
We use MCMC method to extract the promising atoms (see Algorithm \ref{alg1}). According to \cite{George_VariableSelectionGibbsSampling_StatAssoc93} a Markov Chain of $\vect \gamma^{(j)}$'s as in \eqref{Eq:gamma_sequence} obtained by Gibbs sampling converges in distribution to $f(\vect \gamma|\vect y)$. This sequence can be obtained by Algorithm \ref{alg1} and those with highest appearance frequency correspond to the prominent atoms in the dictionary $\mat A$ that can be used for a better reconstruction. %The sequence \eqref{Eq:gamma_sequence} can be obtained by first deriving an auxiliary sequence as suggested by \cite{George_VariableSelectionGibbsSampling_StatAssoc93}:
%\bea
%    \vect x^{(1)}_{t,C_r}~,~\vect \gamma^{(1)}_{t,C_r}~,~\vect x^{(2)}_{t,C_r}~,~\vect \gamma^{(2)}_{t,C_r}~,~    \vect x^{(3)}_{t,C_r}~,~\vect \gamma^{(3)}_{t,C_r}~,~...
%%\label{Eq:AuxSequence}
%\eea
%in which \eqref{Eq:gamma_sequence} is embedded inside.\\
%Initial starting points in Gibbs sampling to obtain the auxiliary sequence are $\vect \gamma^{(0)}  =(1, ,~1,~...,1) $ and $\vect x^{(0)} $ to be the least square estimate of $\vect y = \mat A \vect x$. Other than that, all the other values of $\vect x^{(j)}$ and $\vect \gamma^{(j)}$ are obtained from the following iterative Gibbs sampling framework.
%\begin{enumerate}
%        \item Sample $\vect x^{(j)} $ from $f(\vect x^{(j)} |\vect y  , \vect \gamma^{(j-1)})$ \vspace{-.08in}
%        \item Sample $i^{th}$ element of $\vect \gamma^{(j)} $ denoted by $\gamma^{(j)}_i$ from \\
%         $f(\gamma^{(j)}_i | \vect y,\vect x^{(j)}  , \vect \gamma^{(j)} _{(i)})$ for $i=1...n$\\
%        where $\vect \gamma^{(j)} _{(i)} = (\gamma^{(j)} _{1},...,\gamma^{(j)} _{i-1},\gamma^{(j-1)} _{i+1},...,\gamma^{(j-1)} _{n}),$
%\end{enumerate}
Details of sampling from distributions in Algorithm \ref{alg1} can be found in \cite{Mousavi_OptProb:Techrep13}.\\
After performing Gibbs sampling, prominent dictionary atoms for reconstruction are revealed, then we find the exact contribution of each dictionary atom by solving the following convex optimization problem on the promising subset of dictatory atoms.
%After performing Gibbs sampling, a sequence of $\vect \gamma_{t,C_r}$ vectors for class $C_r$ and task $t$ is obtained. Using these sequences, those elements that are more frequent in \eqref{Eq:gamma_sequence} are identifying the prominent dictionary atoms.
%\begingroup
%\fontsize{13pt}{13pt}\selectfont
\bea
    \vect x_{\vect \gamma}^\ast = \arg \min_{\vect x_{\vect \gamma}} \frac{\sigma^2}{\sigma_n^2} ||\vect y - \mat A_{\vect\gamma} \vect x_{\vect \gamma}||_2^2 + \lambda ||\vect x_{\vect \gamma} ||_2^2
\label{Eq:FinalOpt}
\eea
%\endgroup
where $\mat A_{\vect \gamma}$ is the new dictionary consisting of only contributing atoms and $\vect x_{\vect \gamma} $ is the corresponding coefficient vector for the reduced problem.
Afterward, we place the exact contribution value of each dictionary element into $\vect x$ based on the indicator variable $\vect \gamma$.\\
By doing the same procedure for each task and putting the resulting $\vect x$ and $\vect \gamma$ vectors together, we form $\mat X_{C_r}^\ast$ and $\mat \Gamma_{C_r}^\ast$ matrices for class $C_r$. % An illustration of this procedure for each class is shown in Fig. \ref{Fig:FlowChart}.
Finally, we should do the whole process again in each class and obtain the corresponding ($\mat X_{C_r}^\ast,\mat \Gamma_{C_r}^\ast$) and do the classification based on the value of cost function or residual as was presented in \eqref{Eq:ClassAssignCollaborative1}.% or \eqref{Eq:ClassAssignCollaborative2}.
\begin{algorithm}[t]
\caption{MICHS (per class, per task)}
\label{alg1}
\begin{algorithmic}
\REQUIRE $\mat A, \vect \kappa_t, \vect y $\\
\STATE(1) \textbf{Find contributing atoms:}
Find sequence of $\vect \gamma $ and $\vect x$ \vspace{-1mm}
\bea
    \vect x^{(1)}~,~\vect \gamma^{(1)}~,~\vect x^{(2)}~,~\vect \gamma^{(2)}~,~    \vect x^{(3)}~,~\vect \gamma^{(3)}~,~...
\label{Eq:AuxSequence}
\eea\vspace{-1mm}
\emph{Initialization:} Iteration counter $j=1$, $\vect \gamma^{(0)}  =(1,...,1) $ \\
            ~~~~~~~~~~~~~~~~~~~~~~~~and $\vect x^{(0)} $ to be the least square estimate.
\WHILE{$j<$ MaxIter }
\STATE(1) Sample $\vect x^{(j)} $ from $f(\vect x^{(j)} |\vect y  , \vect \gamma^{(j-1)})$ \vspace{-.08in} \vspace{1mm}
\STATE(2) Sample $i^{th}$ element of $\vect \gamma^{(j)} $ denoted by $\gamma^{(j)}_i$ from \\
         ~~~~~$f(\gamma^{(j)}_i | \vect y,\vect x^{(j)}  , \vect \gamma^{(j)} _{(i)})$ for $i=1...n$\\
         ~~~~~where $\vect \gamma^{(j)} _{(i)} = (\gamma^{(j)} _{1},...,\gamma^{(j)} _{i-1},\gamma^{(j-1)} _{i+1},...,\gamma^{(j-1)} _{n}),$
\ENDWHILE

Extract sequence of $\vect \gamma$ from \eqref{Eq:AuxSequence}: \vspace{-1mm}
\be
    \vect \gamma^{(1)}~~,~~\vect \gamma^{(2)}~~,~~\vect \gamma^{(3)} ~,~...
\label{Eq:gamma_sequence}
\ee \vspace{-1mm}
Take the most frequent $\gamma_i$ s in \eqref{Eq:gamma_sequence} and form $\vect \gamma^\ast$ \\
\STATE(2) \textbf{Find values of contribution} by \eqref{Eq:FinalOpt}\\
\STATE(3) \textbf{Insert Values in $\vect x^\ast$ based on $\vect \gamma^\ast$}
\ENSURE $\vect \gamma ^\ast, \vect x^\ast $
\end{algorithmic}
\end{algorithm}

%% file: section4_v3.tex
\section{Experimental Results}
\label{sec:Results}
\begin{figure}
    \centering
    \includegraphics[scale=0.55]{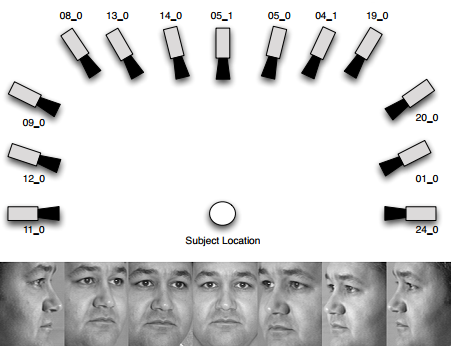}
    \caption{Image acquisition configuration and sample images}
    \label{Fig:Camera}
\end{figure}
\begin{figure}
    \centering
    \includegraphics[scale=0.20]{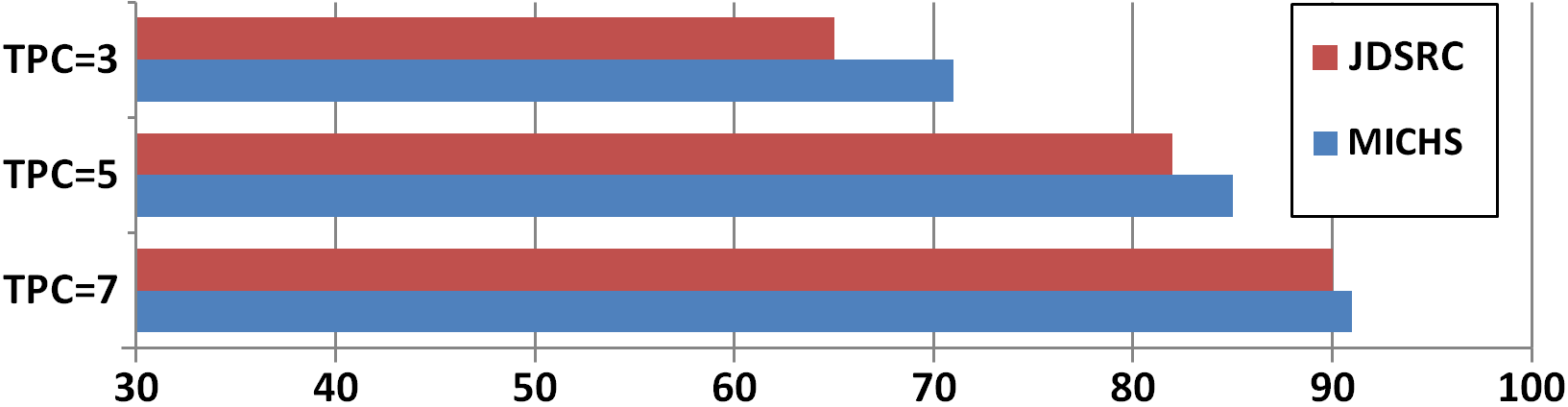}
    \caption{\small {Recognition rates in low training scenario. $C=30$ individuals. Numbers are higher because we reduced the number of classes.}}
    \label{Fig:Result2}
\end{figure}

Multiview face recognition is a multi task image classification problem of significant importance and \emph{Zhang et al.} have done a thorough investigation on this problem in \cite{Zhang:JointDynamicSparseFaceRecognition_pattern12}. In order to validate the performance of our proposed Multitask Image Classification via collaborative Hierarchical Spike and slab priors (MICHS), we present the experimental results of applying MICHS to this problem and compare the results with state-of-the-art algorithms.\\
\textbf{CMU Multi-PIE database:} We conducted the experiments on the CMU Multi-PIE face database \cite{Gross:MultiPIE} which contains a large number of facial images under different illuminations, view points and expressions up to four sessions. There is a collection of cameras at different view angles ($\theta = \{ 0^\circ, \pm15^\circ, \pm30^\circ, \pm45^\circ , \pm60^\circ , \pm75^\circ, \pm90^\circ  \}$) that captures the same scene. Among all the available individuals, we picked $C=129$ subjects that are present in all four sessions of acquisition. Illustration of multiple camera configuration as well as sample images can be found in Fig \ref{Fig:Camera}.\\
We follow the procedure described in \cite{Zhang:JointDynamicSparseFaceRecognition_pattern12} for conducting the experiments and build our training dictionary $\mat A$ by picking training images from session 1 using a subset of available views, i.e. $\theta_{\textit{train}} = \{ 0^\circ, \pm30^\circ , \pm60^\circ , \pm90^\circ \}$. Test images are obtained from all available view angles and from session 2. This is a more realistic scenario that not all the testing poses are available in the dictionary. To generate a test matrix with $T$ views we randomly pick one the $C$ subjects and again randomly pick $T$ different views of that person from $\theta$. We generated two thousand such test samples and compared MICHS \footnote{To benefit from probabilistic nature of the framework a simple and intuitive way of choosing $\mat K$ matrix per class can be found in \cite{Mousavi_OptProb:Techrep13}} performance with classical Mutual Subspace Method (MSM) \cite{Fukui:MSM_Springer05}, Graph-based method in \cite{Kokiopoulou:GraphBasedClassification_Pattern10}, SRC \cite{Wright:SRC_tpami09} combined with majority voting, JSRC method in \cite{Tropp:SimultaneousSparsityGreedyPursuit_SP06} and finally with JDSRC \cite{Zhang:JointDynamicSparseFaceRecognition_pattern12}.\\
To show the effectiveness of our algorithm, first we compared the results for $T=1$ (single task scenario) and then we showed the results where we have $T=3$ different views. These results are shown in Table \ref{Table:Results1} and our method is giving the best performance among all algorithms in both cases. Fig \ref{Fig:Result2} illustrates the results for a scenario with reduced number of classes where we only consider 30 classes out 129 subjects. We compared our result with JDSRC which was shown to be second best after us. We argued in section \ref{sec:CollaborativePriors} that by using class specific priors we can expect to have less sensitivity (slower decay) to insufficiency of number of training samples. This is verified in Fig. \ref{Fig:Result2} with reduced number of Training Per Class (TPC $= 3,5,7$).
\begin{table}
\small
\caption{ {Recognition rate for $C=129$ classes} } % title of Table
\centering
\begin{tabular}{c c c c c c c} % centered columns (4 columns)
\hline\hline %inserts double horizontal lines
View($T$) & MSM & Graph & SRC  & JSRC & JDSRC   & MICHS\\  % inserts table
%heading
\hline % inserts single horizontal line
1 & 36.5 &44.5 &45.0 &45.0 &45.0 & \textbf{51.3}\\ % inserting body of the table
3 & 48.9 &63.4 &59.5 &53.6 &72.0 & \textbf{73.0}\\
\hline %inserts single line
\end{tabular}
\label{Table:Results1} % is used to refer this table in the text
\normalsize
\end{table}
